\documentclass{article}

\usepackage{PRIMEarxiv}
\usepackage{cite}
\usepackage{appendix}
\usepackage{amsmath,amssymb,amsfonts}
\usepackage{algorithmic}
\usepackage{graphicx}
\usepackage{textcomp}
\usepackage{xspace}
\usepackage{url}
\usepackage{multirow}
\usepackage{booktabs}
\usepackage{algorithm} 
\usepackage{algorithmic}
\usepackage{tablefootnote}
\usepackage{caption}
\usepackage{subcaption}
\usepackage{bm}
\usepackage[shortlabels]{enumitem}
\usepackage{natbib}
\usepackage[normalem]{ulem}
\usepackage{xcolor}

\newcommand{\ES}{\textbf{EvS}\xspace}
\newcommand{\SN}{\textbf{SeqNAS}\xspace}

\begin{document}

\title{SeqNAS: Neural Architecture Search for Event Sequence Classification
\thanks{\textit{\underline{Citation}}: 
\textbf{I. Udovichenko et al., "SeqNAS: Neural Architecture Search for Event Sequence Classification," in IEEE Access, doi: 10.1109/ACCESS.2024.3349497.}} 
}

\author{
  Igor Udovichenko, Egor Shvetsov \\
  Skolkovo Institute of Science and Technology \\
  Moscow, Russia \\
  \texttt{\{i.udovichenko, e.shvetsov\}@skoltech.ru} \\
   \And
  Denis Divitsky \\
  Tinkoff \\
  Moscow, Russia \\
  \texttt{divitsky.den2012@yandex.ru} \\
  \And
  Dmitry Osin, Ilya Trofimov \\
  Skolkovo Institute of Science and Technology \\
  Moscow, Russia \\
  \texttt{\{d.osin, ilya.trofimov\}@skoltech.ru} \\
  \And
  Anatoly Glushenko \\
  VTB \\
  Moscow, Russia \\
  \texttt{glushenko@vtb.ru} \\
  \And
  Ivan Sukharev, Dmitry Berestenev \\
  Zvuk \\
  Moscow, Russia \\
 \texttt{\{sukharev-iv, berestnev-da\}@zvuk.com} \\
  \And
  Evgeny Burnaev \\
  Skolkovo Institute of Science and Technology \\
  Moscow, Russia \\
  \texttt{e.burnaev@skoltech.ru} \\
}

\maketitle

\begin{abstract}
Neural Architecture Search (NAS) methods are widely used in various industries to obtain high-quality, task-specific solutions with minimal human intervention. 
Event Sequences (\ES) find widespread use in various industrial applications, including churn prediction, customer segmentation, fraud detection, and fault diagnosis, among others.
Such data consist of categorical and real-valued components with irregular timestamps.
Despite the usefulness of NAS methods, previous approaches only have been applied to other domains: images, texts or time series. Our work addresses this limitation by introducing a novel NAS algorithm --- \SN , specifically designed for event sequence classification. 
We develop a simple yet expressive search space that leverages commonly used building blocks for event sequence classification, including multi-head self attention, convolutions, and recurrent cells. To perform the search, we adopt sequential Bayesian Optimization and utilize previously trained models as an ensemble of teachers to augment knowledge distillation. As a result of our work, we demonstrate that our method surpasses state-of-the-art NAS methods and popular architectures suitable for sequence classification and holds great potential for various industrial applications.
\end{abstract}

\keywords{NAS \and Temporal point processes \and event sequences \and RNN \and transformers \and knowledge distillation \and surrogate models}

\section{Introduction}
\label{motivation}
\textbf{Motivation}. Event Sequences (\ES) with marker and timing information are very common in real-world applications such as medicine~\cite{waring2020automated}, biology~\cite{valeri2023bioautomated}, social medial analysis~\cite{TMPSurvey}, fault diagnosis~~\cite{fault, giurgiu2019explainable}, churn prediction~~\cite{churn1, churn2}, customer segmentation~~\cite{segmentation}, fraud detection~~\cite{fraud} and more. Consequently, there is a demand to model such data. 

In~\cite{giurgiu2019explainable}, the authors focus on predicting failures using telemetry data as a \ES classification task. They emphasize the significant benefits of even predicting a small fraction of these incidents, including improved availability, cost reduction, and avoidance of reactive maintenance.

Based on the  results presented in~\cite{waring2020automated},  applications of Automated Machine Learning (AutoML) in healthcare require additional research and development. The utilization of automated methods in medicine holds great potential for substantially enhancing accuracy, with even minor improvements carrying significant weight in the healthcare sector.

In the field of biology, representing biological datasets as sequences is common. In~\cite{valeri2023bioautomated}, the authors develop an end-to-end automated machine learning tool specifically designed for explaining and designing biological sequences.

In~\cite{boukherouaa2021powering}, the authors explore a wide range of machine learning applications for enhancing  efficiency in the financial sector. These technologies have the potential to automate processes, improve risk assessment and management, and enable more accurate and efficient decision-making. It is important to note that the financial sector, particularly banking, accumulates a significant amount of sequential data based on customer behavior and market events.

\ES classification methods are used in various fields, however, the successful utilization of machine learning still requires substantial effort from human experts, as no algorithm can achieve optimal performance on all possible problems.

\begin{figure}[t]
\centering
\includegraphics[width=\linewidth]{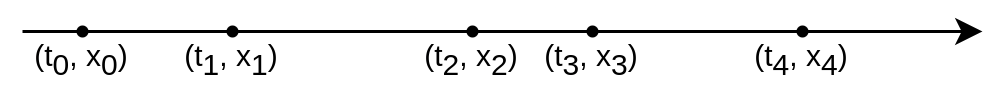}
\caption{An example of the marked temporal event process. Event $i$ occurs at time $t_i$ and is characterized (marked) by the feature vector $x_i$.}
\label{fig:events_sequence}
\end{figure}

Most of machine learning research and applications have been centered around \emph{core domains: text, images, time-series, and speech}. 
However, \ES differ from these well-studied domains in several ways:
\begin{enumerate}[(1)]
    \item events are usually described by both categotical and numerical features,
    \item events in an arbitrary sequence are not uniformly spaced in time (an example can be seen in Figure~\ref{fig:events_sequence}), whereas data in core domains is usually uniformly distributed spatially (image pixels) or temporally (speech signals), and
    \item elements that are close together in space or time have a shared context, and valuable information can be inferred from their neighboring elements. However, this principle does not necessarily apply to \ES since the events may not occur in close proximity to each other in time.
\end{enumerate}

The properties mentioned above can vary significantly across different datasets. As a result, effectively modeling \ES requires the development of task-specific deep learning architectures. This process involves leveraging domain knowledge and can be labor-intensive due to the iterative nature of trial and error. 


Our work aims to develop a NAS procedure specifically designed to effectively handle diverse event sequences. We refer to this approach as \SN (Sequence NAS).

\textbf{Our contributions and results:}
\begin{itemize}
\item{\textbf{Performance}. Our simple yet efficient method \SN shows the superior performance when compared to existing NAS methods and popular architectures that are used for \ES classification.}
\item \textbf{Search Space Design}. We \textit{design a novel search space} of size $\sim 5\times10^6$ possible architectures. The search space contains multi-head self attention, convolutions, and recurrent cells and is tailored to handle event sequence data. To the best of our knowledge, we are the first to develop and analyze such a search space for event sequence datasets.

\item \textbf{Ensemble Of Teachers}. 
Typically, intermediate (suboptimal) architectures obtained during architecture search are thrown away. We propose to utilize them as an ensemble of teachers for subsequent models via knowledge distillation~\cite{distill}.

\item \textbf{Benchmark Datasets}. To advance the development of event sequence classification methods, we have initiated a benchmark for \ES classification by comparing various models and methods. Our study employs six event sequence datasets that were sourced from online competitions held over time or used by other authors. Our work is the first one to carry out a comparison of \ES classification methods for a diverse list of datasets. We make these pre-processed datasets openly available.

\item \textbf{NAS-Bench Event Sequences}.
We present a novel neural architecture dataset comprising 3200 trained architectures, each accompanied by its corresponding scores. This dataset can further facilitate development of predictor-based NAS methods.
\end{itemize}
We provide the source code for the experiments conducted on publicly available datasets, as described in this paper, along with the datasets themselves. \footnote{Our code: \url{https://github.com/On-Point-RND/SeqNAS}}

\section{Related work}
\subsection{Search Space}
The performance of NAS algorithms heavily relies on the search space design, which should exhibit a reasonable degree of flexibility and accommodate established high-performing solutions. A well-designed search space can deliver a satisfactory outcome even with a random search strategy \cite{li2020random}. Therefore, search space design is a primary focus of our work.

Although event sequence data are very common in various applications, the majority of NAS algorithms are designed to solve image classification problems and only rarely extend their search procedures to other core domains. 

Closest to our domain are works exploring text classification and multivariate time-series classification. To evaluate the performance of our procedure for \ES classification, we included methods from both domains in our benchmark and studied their transferability to \ES. These methods are discussed in Sections \ref{NAStext} and \ref{DNNmtv}.

\subsection{NAS for text classification}
\label{NAStext}

The most frequently used building \textbf{blocks} in modern deep neural network (DNN) architectures are: 1) Convolutional;  2) Recurrent; 3) Multi-Head Self-Attention; 4) Pooling Layers; and 5) Identity layers. \textbf{AutoAttend}~~\cite{Autoattend} and \textbf{TextNas}~~\cite{Textnas} have both explored different methods of combining these building blocks as graph nodes to construct a search space. 

In \textbf{TextNas}~~\cite{Textnas}, each node is selected from a pool of blocks described above, and each incoming feature (edge) is selected from a pool of nodes from previous layers.
This work is the closest one to ours.

\textbf{AutoAttend}~~\cite{Autoattend} introduces a NAS procedure to search for attention representations. The main idea is that \textbf{K}-keys, \textbf{V}-values, and \textbf{Q}-queries may originate as features from distinct layers and that different searchable blocks are used to project each incoming feature. Then, incoming features are aggregated in an attention or an addition layer. 
 
In \textbf{NAS-Bench-NLP}~~\cite{klyuchnikov2022bench}, authors search for an Recurrent Neural Network (RNN) structure. The RNN cell is represented as a directed graph in which nodes correspond to specific operations and edges encode their inputs. The authors use four different types of operations in their study, namely: linear layers, element-wise weighted summation, element-wise product, and various activation functions. The operations associated with the nodes and edges are selected during the search process. 

\subsection{DNN for Multivariate Time Series classification}
\label{DNNmtv}

\textbf{Gated-Transformer-"GTN"}~~\cite{Gated} uses two attention blocks instead of one, one to model step-wise and another channel-wise correlation between components of Multivariate Time Series. The performance of this approach for \ES is evaluated in Section \ref{results}. 

In \textbf{ROCKET }~~\cite{ROCKET}, the authors focus on Univariate Time Series (UTS) classification. \textbf{ROCKET} generates feature representations using several randomly initialized convolutional kernels and then trains a linear layer a top of these features. A comparison of \SN and ROCKET can be found in Table~\ref{tab:ucr}. 

\subsection{Search Methods}
\label{Search Methods}
The architecture of a deep neural network (DNN) can be modeled as a directed acyclic graph (DAG), where the nodes represent operations, and the edges denote incoming or outgoing features. In this way, neural architecture search (NAS) is viewed as an algorithms for discovering a task-specific DAG.  

There are various search strategies available, including Bayesian Optimization (BO), Evolutionary Methods, Reinforcement Learning, and Differentiable NAS (DNAS). These methods aim to find the best suitable architecture in the vast space of neural architectures with significantly fewer resources than an exhaustive search requires. 

In ENAS~~\cite{ENAS}, reinforcement learning is utilized to train a super-net, which is an over-parameterized architecture allowing for efficient weight sharing among sub-models. This eliminates the need to train each candidate sub-model from scratch, resulting in a significant reduction in search time. Additionally, DNAS ~\cite{FBNet, Darts, AGD, Trilevel, QuantNAS, mazyavkina2021optimizing} builds upon this idea by representing the over-parameterized super-net as DAG and assigning differentiable importance weights to each edge. The highest edge values determine the selected sub-graph or path. This approach minimizes the need to evaluate multiple models, thereby accelerating the search process.
Knowledge distillation without an ensemble of teachers was applied for NAS in \cite{trofimov2023multi} for image classification.

However, \cite{wang2021rethinking} and \cite{QuantNAS} demonstrated that the optimal architecture is not always selected using DNAS and that the procedure requires various modifications to perform well.
On the other hand, \cite{klyuchnikov2022bench, SNAS} have demonstrated that various BO-based methods perform effectively across varied search spaces, including text classification. One such method, highlighted by the authors, is \textbf{BANANAS}~~\cite{Bananas}, which relies on BO and the neural \textit{Predictor-model}  --- the model designed to predict architecture performance bypassing full train and validation cycle.
\label{sec:precictor_model}
The neural \textit{Predictor-model } is trained on previously queried architectures to score new potential candidates before training them.
Unlike DNAS, \textbf{BANANAS} requires training multiple models. However, we show that previously trained models can be used as a practical advantage. This is discussed in Section \ref{distillation}.

To train a \textit{Predictor-model } on a set of architecture-score tuples, it is necessary to have a procedure for architecture encoding. 
The study~~\cite{white2020study} propose eight architechure encoding schemes categorized into two groups: adjacency matrix-based and path-based.
The authors evaluate the performance of each encoding scheme for different NAS subroutines: \textit{Predictor-model } training, architecture perturbation and random architecture sampling.
They show that no encoding scheme performs well across all subroutines, but the path-based encoding outperforms the adjacency matrix-based one on the task of training the \textit{Predictor-model }.

\subsection{NAS Benchmarks}
The NAS-Bench series of benchmarks ~\cite{li2021hw, nas-bench-101, dong2020bench, siems2020bench, klyuchnikov2022bench, mehrotra2020bench} have made significant contributions to the advancement of scientific research in neural architecture search (NAS). These benchmarks aim to establish a standardized measurement procedure and provide datasets for easy comparison and reproducibility in NAS research. They include datasets of trained architectures and their corresponding scores, along with detailed discussions on the characteristics and performance of various NAS algorithms. However, these benchmarks have not yet explored the domain of event sequence. In our work, we extend the NAS-Bench series with NAS-Bench Event Sequences, a dataset of architectures specifically designed to model \ES.

\subsection{Temporal Point Processes  modeling}
Recently, different neural architectures and approaches were used to model \ES as Temporal Point Processes (TPP)~~\cite{TMPSurvey}. These data exhibit complex short-term and long-term temporal dependencies. Existing methods heavily rely on Recurrent Neural Networks (RNNs) due to the sequential nature of event sequences \cite{du2016recurrent}. However, RNN units are not effective in capturing long-term dependencies. On the other hand, transformer and convolutional-based models are capable of handling long-term dependencies, but they assume a uniform temporal distribution. To address these challenges, authors in \cite{zuo2020transformerhawkes} propose a transformer-based architecture that models the dynamics of temporal point processes using a continuous conditional events intensity function. Additionally, in \cite{romero2021ckconv}, a long convolutional kernel with weights, conditioned on event samples intensity, is prpoposed. This parameterization enables the handling of non-uniformly sampled and irregularly-sampled datasets.

\begin{figure}[h]
\centering
\includegraphics[scale=0.8]{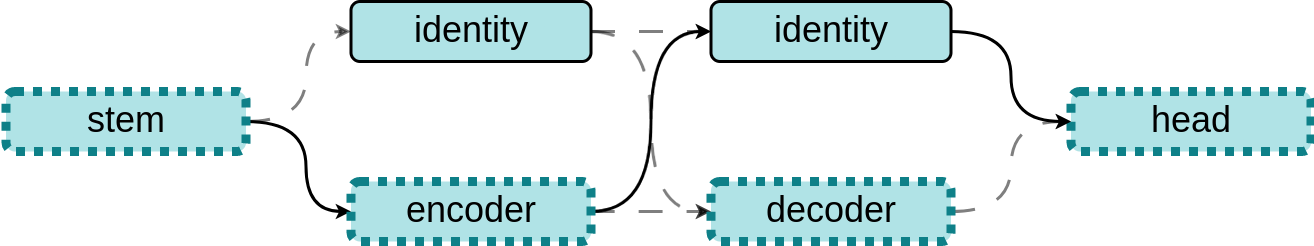}
\caption{The general layout of our search space. Dotted borders indicate that blocks contain searchable operations. Dashed lines indicate that connections between nodes are searchable. The solid line is an example of selected architecture.}
\label{fig:ss_gen_struct}
\end{figure}

\section{Methodology}
\subsection{Search Space design}\label{search space design}
\begin{figure}[h]
    \centering
    \includegraphics[width=0.5\linewidth]{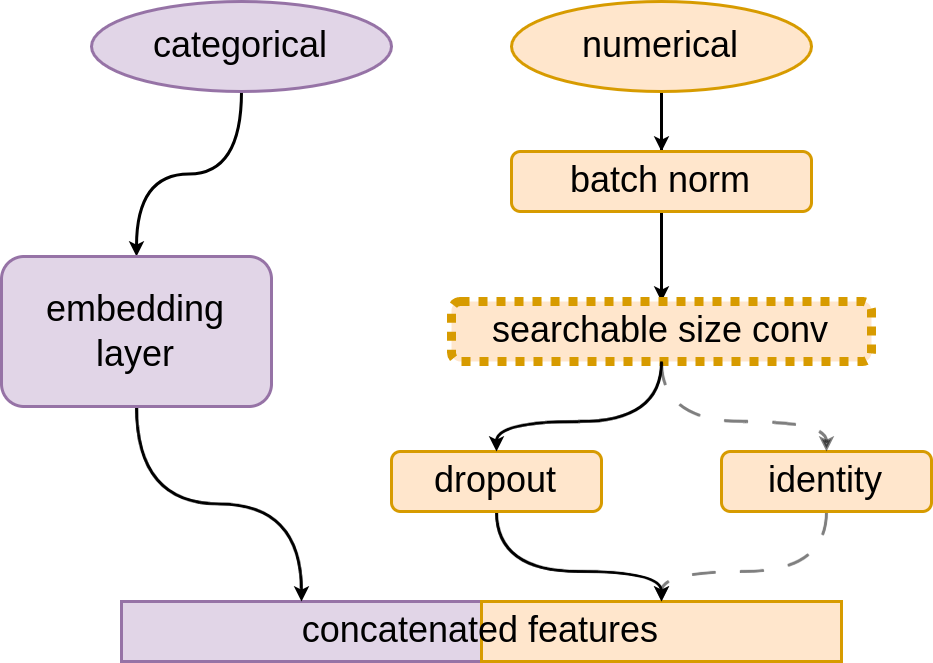}
    \caption{Searchable part of \textbf{Stem block} is depicted with dashed and dotted lines. Convolutional layers with different kernels and the presence of dropout are selected at each search step. A solid line is an example of a selected path.}
    \label{fig:ss_stem}
\end{figure}

\begin{figure}[h]
\centering
\includegraphics[width=\linewidth]{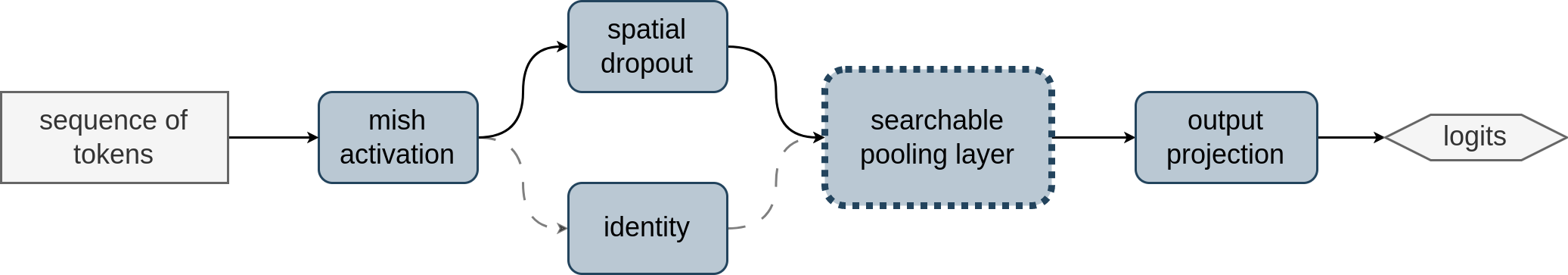}
\caption{There are two searchable pooling layers in \textbf{Head Block}: Max pooling and Average pooling. The type of a pooling layer and the presence or absence of spatial dropout are determined by the search procedure. A solid line is an example of a selected path.}
\label{fig:ss_head}
\end{figure}

The general layout of all the blocks in our search space is illustrated in Figure~\ref{fig:ss_gen_struct}.
There are four main blocks in the search space: 
\begin{itemize}
    \item \textbf{Stem} \textit{(always present)} has a searchable structure depicted in Figure~\ref{fig:ss_stem}. 
    \item \textbf{Encoder} \textit{(optional)} is searchable with multiple layers, whose structure is depicted in Figure~\ref{fig:ss_layer}.
    \item \textbf{Decoder} \textit{(optional)}  has a searchable number of layers.  
    \item \textbf{Head} \textit{(always present)} has a searchable structure depicted in Figure~\ref{fig:ss_head}.
\end{itemize}
Here, the term \textit{optional} denotes that the presence of a particular block is determined during the search procedure. For instance, the minimal architecture would consist only of the \textbf{Stem} and \textbf{Head} blocks.
Now we describe earch architecture block in more details.

\subsubsection{Stem} The Stem fuses categorical and numerical features from input data into one vector as depicted in Figure~\ref{fig:ss_stem}. 

The Stem pipeline is threefold: 
\begin{itemize}
\item Categorical features are encoded using an embedding layer, and the size of the embedding is automatically determined by the formula: $\min(600, \mathrm{round}(1.6 \times N^{0.56}))$. 
Where $N$ is a sequence length. 
\item Numerical features are processed using BatchNorm~\cite{santurkar2018does}; afterwards, convolution with a searchable kernel size is applied to each numerical input along the temporal dimension, and optionally, dropout may be applied after convolution. 
\item Finally, all embeddings of all types are concatenated along the feature dimension to obtain the input sequence. 
\end{itemize}

\subsubsection{Encoder} 
Encoder brings the most variability into the search space. 
Common operations are available in the encoder:  Multi-Head Self-Attention (MHA)~\cite{vaswani2017attention}, Gated Reccurent Unit (GRU)~\cite{chung2014empirical}, and Convolution. 
Each of these operations entails different assumptions about the nature of the data: RNN units assume a sequential nature of the data, convolutional layers are effective at capturing temporally local correlations, and transformers excel at capturing long-term dependencies throughout the entire sequence.

\begin{figure}[h]
    \centering
    \includegraphics[width=0.4\linewidth]{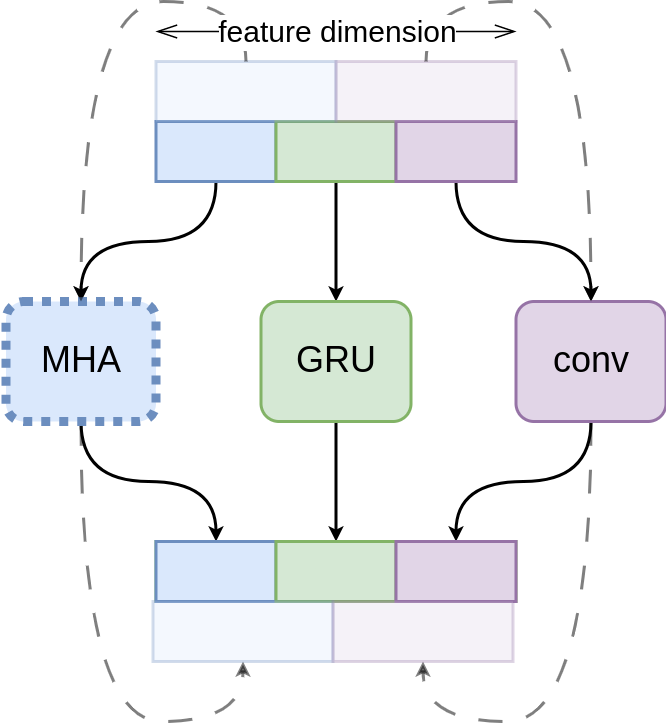}
    \caption{Encoder Layer with searchable \textbf{MHA}, \textbf{GRU} and \textbf{conv} operations. A combination of one, two, or three operations can be selected during each search step. Different combinations are selected on different layers. Incoming features are divided into several selected operations. An example combination with \textbf{MHA}, \textbf{GRU} and \textbf{conv} operations is depicted with solid lines, and an example combination with \textbf{MHA} and \textbf{conv} operations is depicted with dashed lines. Dotted border around \textbf{MHA} indicates that it has a searchable number of heads.}
    \label{fig:ss_layer}
\end{figure}

Encoder has both a searchable number of layers and the operations within each layer.
Input of each Encoder layer is divided into one to three blocks along the feature dimension, which can be processed using one of six potential operations, such as MHA, GRU, or Convolution. 
The number of heads in MHA is searchable and is chosen from the set $\{1, 2, 4, 8\}$.
In total it provides up to 19 variations for a single Encoder layer. 
It is worth noting that each layer has a distinct set of operations. The outputs from each block are concatenated and sent to the next layer within the Encoder. This structure is illustrated in Figure~\ref{fig:ss_layer}. 
The encoder may have three possible values for the number of layers (1, 2, or 4), resulting in approximately $130\times 10^3$ possible Encoder variations.
\subsubsection{Temporal encoding}
The traditional positional encoding uses the token position in the sequence to obtain the embedding~\cite{vaswani2017attention}.
However, in the case of non-uniformly spaced event sequences, this traditional positional embedding is inadequate for capturing the relative arrangement of events. Therefore, in line with the approach proposed in Transformer Hawkes process \cite{zuo2020transformerhawkes}, we utilize temporal encoding to address this limitation.
\begin{align*}
    TE(t, 2i) &= \sin\left( t / 10000^{2i / d}\right), \\
    TE(t, 2i+1) &= \cos\left( t / 10000^{2i / d}\right),
\end{align*}
where $t$ is an event time normalized to $[0, 1]$, $i$ --- dimension, and $d$ is the embedding size.

\subsubsection{Decoder} 
In the decoder block, we utilize a standard transformer architecture with a searchable number of heads and layers without recurrent or convolutional layers used in the encoder. We adopt the transformer variant proposed in \cite{wang2022foundation} with Sub-LayerNorm and same weights initialization.

\subsubsection{Head} 
Head aggregates the sequence of feature vectors to produce the final classification as depicted in~Figure~\ref{fig:ss_head}. First, optional spatial dropout operation is performed ~\cite{tompson2015efficient}. Next, the sequence of tokens is aggregated into a single feature vector, aggregation is also a searchable operation consisting from either the maximum operation, averaging, or a combination of both. Finally, the feature vector is projected to obtain the final logits.

\subsection{The Architecture Vectorizer}
To extract the architecture features for a \textit{Predictor-model } - the model designed to predict architecture performance, we focus on a group of path-based encoders described in~~\cite{white2020study}. 
According to~\cite{white2020study} Path-based encoders outperform the adjacency matrix-based ones for the \textit{Predictor-model } setting. Our encoding is done as follows. A binary variable is assigned to each block, layer and the particular operation in each layer.
If the block or layer is not involved in the architecture, the corresponding binary variable, and all variables responsible for the operations inside the block or layer are set to zero.
Finally, all variables are concatenated to obtain the feature vector. For simplicity we call our architecture encoding scheme --- \textbf{AVec}. 

\subsection{The search procedure}
\label{our_search}
With the reasoning presented in Section~\ref{Search Methods}, we have opted to use a bayesian optimization similar to~\cite{Bananas}. However, instead of an ensemble of DNNs we employed \textbf{CatBoost} ~\cite{CatUncert} to obtain predictions and corresponding uncertainty estimates. This alteration from~\cite{Bananas} has enabled us to leverage the benefits that \textbf{CatBoost} offers over  the ensemble of DNNs, a more  precise uncertainty estimation following a theoretically justified approach~\cite{CatUncert}. We analyze this choice in Section~\ref{sec:cat_vs_dnn}.
The search procedure is outlined in Algorithm~\ref{alg:search}. There are \textbf{three main components of the search process}: 1)~Architecture vectorizer --- \textbf{AVec}, 2)~Score prediction and uncertainty estimation --- \textit{Predictor-model} as CatBoost, 3)~Candidate selection --- \textit{Thompson sampling}.
    
Initially, a set of $N_{init}$ architectures is randomly sampled from the search space $A$. 
After training all of them, actual performance scores are obtained for each architecture. 
Next, the $Spredictor$ is trained using the architecture features and actual scores.
Architecture features are obtained with our architecture vectorizer --- \textbf{AVec}.
Then, the trained $Predictor-model  $ is used to estimate scores and associated uncertainties for new randomly sampled $N_{iter}$  architectures. 
It is crucial to balance the exploration-exploitation trade-off during the search process.
To achieve the balance we use Thompson sampling with estimated scores and uncertainties. 
$L_{candidates}$ architectures are sampled for further trainig.
These steps are repeated until the allocated budget is met as described in Algorithm~\ref{alg:search}.

For further technical details on each parameter, Section~\ref{tech_details} provides a detailed explanation.

\subsection{Ensemble Of Teachers}
\label{distillation}
Our search procedure involves training many models. By combining these models as an ensemble of teachers, we are able to leverage the benefits of this approach. In ~\cite{BadModels}, authors demonstrated that a weak teacher ensemble could lead to improved student performance. This observation allows us to construct an ensemble of best-performing models at a current search step. We use an average of different models predictions as a teacher model. Before enabling distillation loss, we train a total of $30$ architectures. On each search iteration all models predictions for all training examples are being cached to avoid computational complexity. At each new iteration, we update members of the ensemble by selecting $top K$ best performing models from our cached predictions. We use ensembles of three models in most of our experiments.

To compute the model-to-model loss, we opted for Mean Squared Error (MSE) instead of Kullback–Leibler divergence. This choice allows us to avoid using additional temperature hyper-parameter~\cite{kim2021comparing}.

\begin{algorithm}[h]
	\caption{$Predictor-model  $ --- model score predictor with parameters $\theta$, $N_{init}$ ---  initial number of architectures to train, $N_{iter}$ --- number of architectures to sample during each iteration, $L_{candidates}$ --- number of architectures to train during each iteration, $Ensemble$ --- ensmbling function, $M$ --- number of iterations, $\hat{X}, S$ --- predicted scores and corresponding uncertainties.} 
        \label{alg:search}
	\begin{algorithmic}[1]
	\STATE $K_{1} \leftarrow Sample(N_{init})$, sample random architectures from the search space.
    \STATE $TrainedArches\leftarrow Train(K_{1})$. Train all architectures in $K_{1}$ and obtain their scores, $X$ is a set of scores from all trained models, $X_i$ are scores for a current iteration.
    \STATE $ArchFeatures \leftarrow AVec(TrainedArches)$, encode architectures into features .
    \STATE $\underset{\theta}{\text{min}} (MAE(Predictor-model  (ArchFeatures;\theta), X))$, train a score predictor model   .  
		\FOR {$i=1,2,\ldots, M$}
		\STATE $K_{1+i} \leftarrow Sample(N_{iter})$, sample random architectures from the search space such that $K_{i+1} \cap TrainedArches = \emptyset$.
            \STATE $\hat{X}, S = Predictor-model  (K_{i+1}; \theta)$, predict scores and score uncertainties .
            \STATE Select $L_{candidates}$ architectures from $K_{1+i}$ with \textbf{Thomson sampling} using obtained uncertainties $S$.
            \STATE $T \leftarrow topK(TrainedArches)$, select the best performing teacher models from already trained models and obtain an ensemble of teachers $Ensemble(T)$.
            \STATE Train all models in $L_{candidates}$ with distillation loss and $Ensemble(T)$ and obtain actual scores $X_i$.
            \STATE $TrainedArches \leftarrow TrainedArches\cup L_{candidates}$.
            \STATE $X \leftarrow X\cup X_i$.
            \STATE $ArchFeatures = AVec(TrainedArches)$, encode architectures into features .
            \STATE Update a score predictor model $\theta \leftarrow \arg \min_\theta \mathrm{L}(\theta)$, where
            $\mathrm{L}(\theta) = MAE(Predictor-model  (ArchFeatures;\theta), X)$.
		\ENDFOR
    \STATE Select the best architecture from $TrainedArches$ according to some performance metric.
	\end{algorithmic} 
\end{algorithm}

\begin{table*}[ht]
\centering
\caption{Statistics of sequential datasets used for our analysis.} 
\label{table:datasets}
\begin{tabular}{lrrrrrr}
\toprule
 & Amex & ABank & VBank & AGE & RBchurn & Taobao \\
\midrule
Number of observations / Transactions & 5531451 & 270450066 & 7636113 & 26450577 & 490513 & 7893060 \\
An average number of observations per user & 12 & 280 & 897 & 881 & 98 & 806 \\
STD for the number of observations per user  & 2.68 & 270 & 371 & 124 & 78 & 1042 \\
Number of users & 458913 & 963812 & 8509 & 30000 & 5000 & 9791 \\
Number of classes & 2 & 2 & 2 & 4 & 2 & 2 \\
Number of categorical features & 11 & 16 & 2 & 1 & 1 & 2 \\
Number of real-valued features & 177 & 3 & 1 & 1 & 5 & 0 \\
Class balance & 0.26 & 0.01 & 0.75 & 0.25 (equal) & 0.65 & 0.6 \\
Target & Default & Default & Education & Age group & Churn & Payment  \\
\bottomrule
\end{tabular}
\end{table*}

\begingroup
\renewcommand{\arraystretch}{1.2}
\begin{table*}[ht]
    \caption{Comparison of our method with two NAS procedures 1)AutoAttend~\cite{Autoattend}, 2)TextNAS~\cite{Textnas} and four fixed architectures   3)Gated Transformer Networks~\cite{Gated}, and baseline models such as 4) Fixed Transformer, 5)GRU, 6)LSTM.  We report MEAN and STD of the 3 best models found, for both HPO and NAS procedures.  We mark the \textbf{First} and the \textcolor{cyan}{Second} best performing models as highlighted in this text.}
    \label{tab:main_res}
    \centering
    \resizebox{\textwidth}{!}{%
    \begin{tabular}{llccccccc}
    \toprule
     \multirow{3}[3]{*}{Dataset} & Model Search Space   & SeqNAS  & AutoAttend  &   TextNAS & GTN &  Fixed TF &      GRU  &       LSTM \\
     \cmidrule(lr){2-9}
     & Metric / Search Method & Our & Context-Aware  &  ENAS & HPO &  HPO & HPO & HPO \\
    & & & Weight Sharing & &  & &  & \\
    \midrule
    AmEx &       Custom\tablefootnote{\url{https://www.kaggle.com/competitions/amex-default-prediction/overview/evaluation}} &        \bm{$0.7911 \pm 0.0004$} &   $0.6170 \pm 0.0033$ &  $0.7818 \pm 0.002$ &       $0.7717 \pm 0.006$  &            \textcolor{cyan}{$0.7850 \pm 0.0002$} &   $0.7718 \pm 0.0005$ &  $0.7709 \pm 0.0005$ \\
    ABank &      ROC-AUC &       \bm{$0.7963 \pm 0.0014$} &   $0.6827 \pm 0.0160$ &  $0.7653 \pm 0.002$ &  $0.7462 \pm 0.001$  & \color{cyan}{$0.7747 \pm 0.0011$} &  $0.7699 \pm 0.0002$ &    $0.7451 \pm 0.0032$ \\
    VBank &      ROC-AUC &      \bm{$0.8032 \pm 0.0022$} & $0.6533 \pm 0.0408$ &  $0.7951 \pm 0.001$ & $0.7362 \pm 0.001$ & $0.7883 \pm 0.0013$ & \color{cyan}{$0.7980 \pm 0.0008$} & $0.7704 \pm 0.0012$ \\
    RBchurn & ROC-AUC & \bm{$0.8525 \pm 0.0033$} & $0.7345 \pm 0.0028$ & $0.7936 \pm 0.002$ & $0.7701 \pm 0.003$ & $0.8170 \pm 0.0012$ & \textcolor{cyan}{$0.8300 \pm 0.002$} & $0.8090 \pm 0.0027$ \\
    AGE & Accuracy & \bm{$0.6445 \pm 0.0018$} & $0.6251 \pm 0.0013$ & $0.6016 \pm 0.003$ & $0.5363 \pm 0.019$ & $0.6170 \pm 0.001$ & \color{cyan}{$0.6300 \pm 0.001$} & $0.5920 \pm 0.0010$ \\
    TaoBao & ROC-AUC & \bm{$0.7138 \pm 0.0007$} & $0.6352 \pm 0.0023$ & $0.7079 \pm 0.002$ & $0.6713 \pm 0.001$ & \color{cyan}{$0.7107 \pm 0.0011$} & $0.7100 \pm 0.0004$ & $0.6680 \pm 0.0008$ \\
    \bottomrule
    \end{tabular}
    }
\end{table*}
\endgroup

\section{Experiments and results}

\subsection{Datasets}

We utilize six publicly available datasets consisting of event sequences sourced from different data science competitions and prior studies.
These sequential datasets were carefully selected to include both categorical and real-valued features.
In each dataset, a sequence of events is provided as an input to predict a categorical target, making it a classification task.  
Detailed statistics and target regarding each dataset can be found in Table~\ref{table:datasets}.

\textbf{Bank transaction data:} \textbf{VBank}, \textbf{AmEx}, \textbf{AGE}, \textbf{RBchurn} and \textbf{ABank} datasets consist of card transactions, financial records, and other user-related data. Leveraging these datasets, we utilize event sequences to predict specific targets such as default events, churn, user higher education, age and etc. Mainly each transaction is characterized by its date, type, amount, and Merchant Category Code.

\textbf{Taobao:} Taobao dataset is a subset of the Taobao APP user behavior data, comprising millions of items recorded over one month. The dataset is organized in a user-item interaction format, consisting of user ID, item ID, category ID, behavior type, and timestamp.

To suit the context of our task, we preprocess the dataset by excluding the item ID for simplicity. Additionally, we merge all categories that appear less than 500 times in the dataset into a single category. This preprocessing step allows us to reduce the number of unique categories from 8,900 to 1,900. We focus on the client's behavior within a 7-day window to predict whether they will make a payment in the following 7 days. 

No manual feature generation or preprocessing was conducted on most of the datasets, except the Taobao and AmEx datasets. In the case of the AmEx dataset, we utilized a cleaner version obtained from the Kaggle competition platform. 

To create train, test, and validation sets, we performed a random split for each dataset. The split ratios were set to 0.6 for the training set, 0.2 for the test set, and 0.2 for the validation set, based on the total sample size. Sequences shorter than specified in Table~\ref{table:datasets} were padded with zeros; for sequences longer than specified, we took $N$ last events, where $N$ is the sequence length specified.  

\subsection{Methods}
We compared our results with two NAS approaches, namely \textbf{AutoAttend}~\cite{Autoattend} and \textbf{TextNAS}~\cite{Textnas}. However, we had to make some modifications to adapt these methods for the \ES domain. These modifications are described in Appendix~\ref{tech_details}. Furthermore, we also compared our approach with fixed architectures such as GRU, LSTM, and Transformer. To ensure optimal performance, we performed Hyper-Parameter Optimization (HPO) for all fixed architectures. The details of the HPO can be found in Appendix~\ref{tech_details}.

\subsection{Results}\label{results}
Our main results are presented in Table~\ref{tab:main_res}. 
\SN outperforms all other methods. The second place is shared by fixed architectures: \textbf{Transformer} and \textbf{GRU}. We further analyze importance of \textbf{MHA} and \textbf{RNN} as building blocks of our search space in Section~\ref{sec:blocks_importance} and show that all blocks are complementary to each other.

Our experiments show that search spaces from related domains, such as text, do not always transfer well to \ES and sometimes underperform even simple models such as RNN. 

A potential disadvantage of \SN is its longer training time compared to other approaches, as shown in Table~\ref{tab:costs}. However, as discussed in Section~\ref{motivation}, it is still a reasonable time complexity given significant gains for various applications.

\subsection{Ablation studies}

\subsubsection{Ensemble Of Teachers.}

  \begin{figure}
  \centering
		\includegraphics[width=0.5\linewidth]{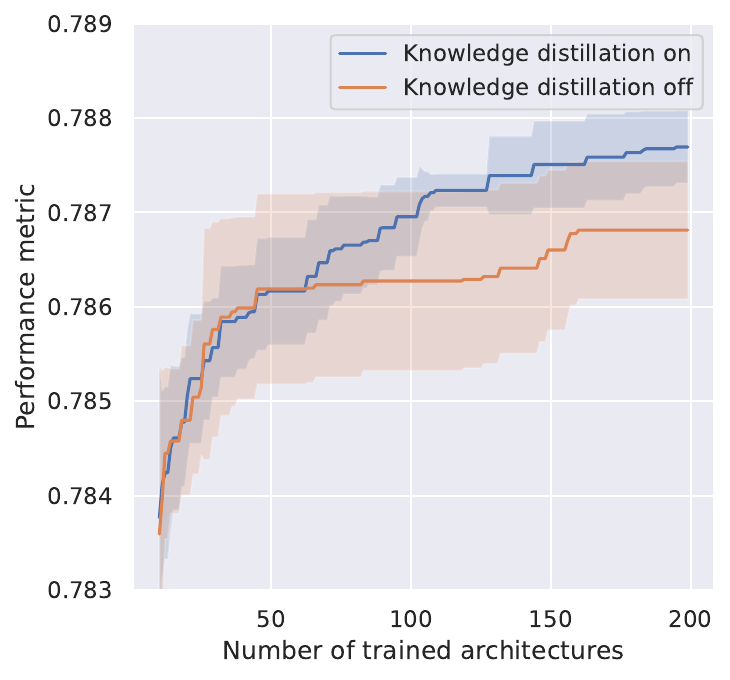}
\caption{Search performance of \SN with and without KD  on AmEx dataset over a number of trained architectures, for 200 architectures in total. Results are averaged over 3 best performing models as a sliding window. Performance is measured with a metric specified in Table~\ref{tab:main_res} for AmEx dataset. KD is employed after 30 models were trained. It can be seen that lines start to diverge approximately after 60th iteration.}
		\label{fig:kd}
  \end{figure}

 \begin{figure}
 \centering
		\includegraphics[width=0.5\linewidth]{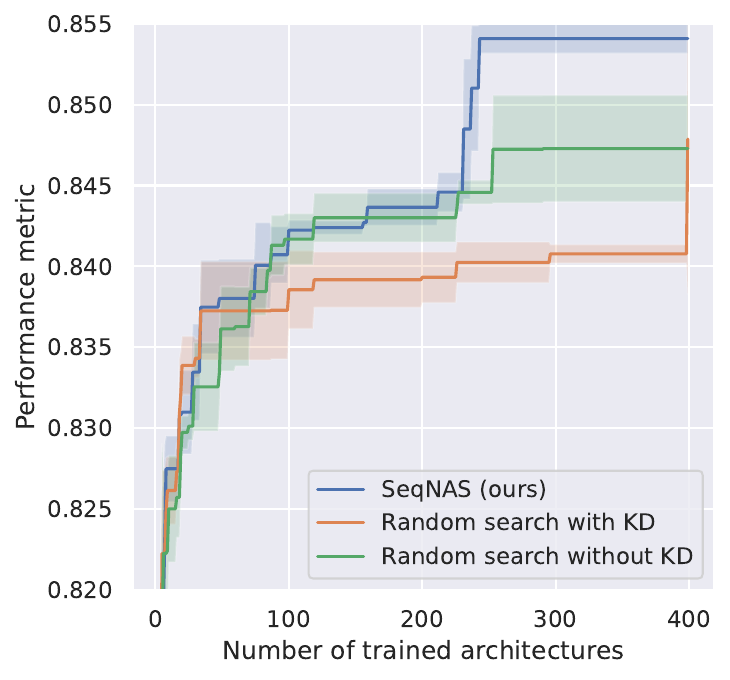}
\caption{Comparison of \SN, Random search with KD and Random search without KD on RBchurn dataset over a number of trained architectures, for 400 architectures in total. Results are averaged over 3 best performing models as a sliding window. Performance is measured with a metric specified in Table~\ref{tab:main_res} for RBchurn dataset. KD is employed after 30 models were trained.}
		\label{fig:kd_random}
  \end{figure}

In Figures \ref{fig:kd} and \ref{fig:kd_random} we demonstrate the results of the search procedure with and without Knowledge Distillation (KD) on two datasets,  AmEx and RBchurn correspondingly. We show the average scores of the top 3 models over the search steps. It can be seen that KD significantly improves performance metrics for both datasets. The same observations can be seen for other datasets in Table~\ref{table:kd}, where we demonstrate the final metrics for the single best architecture using a random search procedure.

We experimented with different approaches to (1) select diverse teachers and (2) combine their predictions into an ensemble. We found that averaging the predictions of 3 best performing models resulted in the best performance. During training, we had access to hundreds of trained models, and we believe there are further opportunities for improvement and potential applications in this direction.

\subsubsection{CatBoost vs. DNN predictior}
\label{sec:cat_vs_dnn}
\begin{figure}[t]
\centering
\includegraphics[width=.5\linewidth]{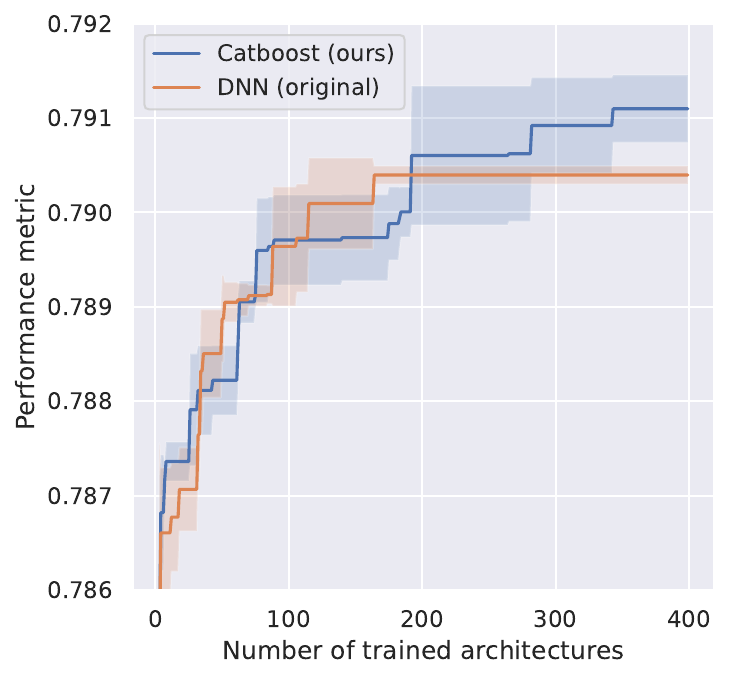}
\caption{Comparison of different predictors on AmEx dataset: \textbf{CatBoost}~~\cite{CatBoost} and an ensemble of DNNs originally proposed in \textbf{BANANAS}~~\cite{Bananas}. Results are averaged over 3 best performing models as a sliding window. Performance is measured with a metric specified in Table~\ref{tab:main_res} for AmEx dataset.}
\label{fig:predictors}
\end{figure}

We evaluate two types of models for architecture scoring and uncertainty estimation ($Predictor-model  $) in Figure~\ref{fig:predictors}. The results demonstrate that the model based on CatBoost outperforms the one based on an ensemble of DNNs with slightly superior performance. We used an ensemble of eight models.

\begin{table}
		\caption{Effect of knowledge  distillation\\ on search score using random search.}
		\label{table:kd}
		\centering
		\begin{tabular}{lllc}
			\toprule
			  Dataset  & Metric  & KD & Score \\
			\midrule
                \multirow{2}{*}{VBank} & \multirow{2}{*}{ROC-AUC} & on  & \bf{0.7997}  \\
                                       & & off & 0.7986       \\
                \multirow{2}{*}{AmEx}  & \multirow{2}{*}{Custom} & on  & \bf{0.7889}  \\
                                       & & off & 0.7876     \\
                \multirow{2}{*}{Insect}& \multirow{2}{*}{Accuracy}  & on  & \bf{0.6164}  \\
                                       & & off & 0.6122      \\
			\bottomrule
		\end{tabular}
  \end{table}

\begin{figure}[t]
\centering
\includegraphics[width=.5\linewidth]{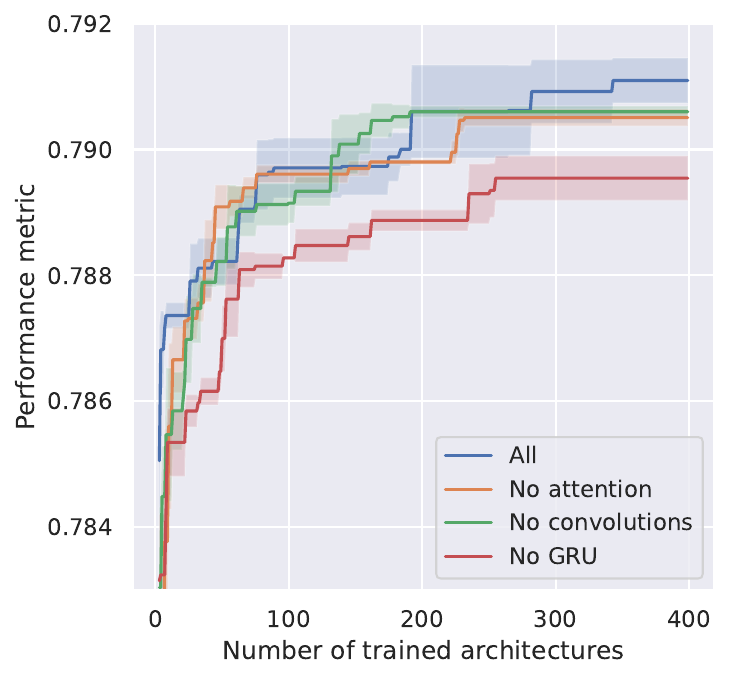}
\caption{Search performance of \SN without different blocks in Encoder layer and with all blocks included --- $ALL$ on AmEx dataset.  Results are averaged over 3 best performing models as a sliding window. Performance is measured with a metric specified in Table~\ref{tab:main_res} for AmEx dataset. We see that different types of operations complement each other.}
\label{fig:sp_ablation_nas}
\end{figure}

\subsubsection{Encoder blocks importance: MHA, GRU, or Convolutions}
\label{sec:blocks_importance}

To better understand the roles of MHA, GRU, and Convolutions in the encoder layers, we conducted a search procedure where we removed one of these blocks at a time. As shown in Figure~\ref{fig:sp_ablation_nas}, models without GRU block exhibited a significant drop in performance compared to those with all three blocks present. Unsurprisingly, these results can be explained by the good performance of GRU model alone presented in Table~\ref{tab:main_res}.   Nonetheless, it is worth noting that the relative importance of each block varies depending on the dataset used.

\begin{figure}
\centering
\includegraphics[width=.5\linewidth]{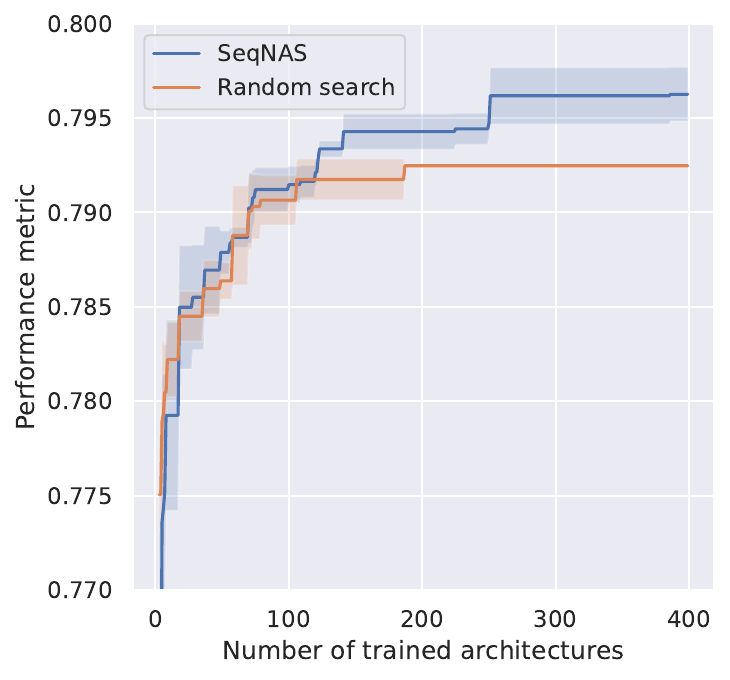}
\caption{Search performance of \SN and Random search with KD.  Results are averaged over 3 best performing models as a sliding window. Performance is measured with a metric specified in Table~\ref{tab:main_res} for ABank dataset. Both \SN and Random search use the ensemble of teachers.}
\label{fig:sp_ablation_nas_random}
\end{figure}

\subsubsection{Random search vs \SN}
In Figures \ref{fig:sp_ablation_nas_random} and \ref{fig:kd_random}, we compare \SN and Random search procedure for two datasets, AmEx and RBchurn correspondingly. In our settings, we first train $100$ randomly sampled architectures and then fit $Predictor-model  $ to score new candidates. It can be seen that \SN starts to outperform Random search approximately after $100$ steps for both datasets. 

\subsection{NAS-Bench Event Sequences}
Analogous to the precomputed NAS benchmarks~\cite{nas-bench-101, dong2020bench, klyuchnikov2022bench}, we release a dataset with trained and evaluated architectures.
The details can be found in Appendix~\ref{sec:nas-bench-es}.

\section{Discussion}
\begin{itemize}
\item The search space of \textbf{TextNAS} contains same operations as \SN. However, \textbf{TextNAS} differs from \SN in terms of graph topology and search procedure. \textbf{TextNAS} performs worse than both \SN and fixed architectures with HPO, raising questions about the impact of search space design or search method, such as ENAS.

\item It's important to note that \SN does not outperform \textbf{ROCKET} for univariate time series classification. Many recent methods that perform well on UCR datasets utilize specific convolutional operations or fixed feature generation. Thus, our search space could benefit from incorporating new searchable operations proposed in various works for UTS classification.
\item There are potential improvements that can be made based on TPP modeling works, such as better estimation of event densities~\cite{zuo2020transformerhawkes} or the use of temporally parameterized long convolutions~\cite{romero2021ckconv}.
\item While our results highlight the importance of the GRU unit in \ES modeling, this observation may be specific to our datasets. It's possible that with larger datasets, Transformer-based blocks may offer better performance. Different training strategies, like applying causal masks, may also help improve Transformer performance.
\item Currently, our method may produce over-parameterized models without considering hardware constraints. Further improvement can be done to develop a more computationally efficient approach.
\item The literature currently lacks a NAS procedure that features a broad search space suitable for a wide variety of tasks, referred to as the \textbf{Universal Search Space.} Moreover, we see that the generalization ability of existing search methods is limited even across similar domains.
\end{itemize}

\section{Conclusions}
In this paper, we introduce \SN, a novel method for automatically searching neural architectures specifically designed for \ES data. Our approach outperforms other NAS methods and standard architectures with hyper-parameter optimization in the \ES domain. We demonstrate the versatility of our method by applying it to various datasets. To the best of our knowledge, our work represents the first extensive exploration of NAS for \ES.

We show that in our search space different types of operations complement each other, leading to the discovery of improved architectures. There is no architecture which performs better without one of the operations: MHA, RNN unit, or convolution.

Our approach combines knowledge distillation with sequential Bayesian Optimization to achieve significant performance improvements in a computationally efficient way. 

Additionally, we establish a benchmark for \ES classification by comparing different models and techniques. This benchmark can serve as a valuable resource for researchers looking to advance the field of \ES classification. 

We release the \textbf{NAS-BENCH Event Sequences} dataset, which includes architectures and corresponding scores, to support research on predictor-based NAS methods.


\section*{Acknowledgment}
The work was supported by the Analytical center under the RF Government (subsidy agreement 000000D730321P5Q0002, Grant No. 70-2021-00145 02.11.2021).


\bibliographystyle{unsrt}
\bibliography{biblio.bib}

\appendix
\section{Techical details}\label{tech_details}

In \textbf{SeqNAS}, for all of our datasets, we used the following hyper-parameters parameters: $N_{init}=100$, $N_{iter}=100$, $M=40$ and $L_{candidates}=15$.  For more details regarding each dataset, please refer to our repository. 

To evaluate various NAS methods, including \textbf{TextNas}, \textbf{AutoAttend}, and \textbf{GTN}, we used the hyper-parameters and search procedures outlined in their original papers. 

For \textbf{TextNas} and \textbf{AutoAttend}, we incorporated  ur Stem blocks to combine real-valued and categorical features. Similarly, for \textbf{GTN}, we added embedding layers for categorical features but chose not to utilize convolutional layers for real-valued features, as they were not utilized in the original paper.

For fixed transformer architecture, we used a simple model with two MHA layers in both Encoder and Decoder, with 8 heads in each.
LSTM and GRU models consisted of only one RNN layer, with Stem and Head blocks. Fixed models were optimized using hyper-parameters optimization.

For all models, we used the identical fixed structure for  Stem and Head blocks described in our architecture. In the Stem block, we did not use dropout and set fixed convolutional layer kernel size to $3 \times 3$. For Head, we fixed the spatial dropout rate at 0.3, and for pooling we used Max pooling.

\label{HPO}
\textbf{For HPO}, we employed Optuna~\cite{Optuna} and optimized the following hyper-parameters for 30 iterations: 1) batch size, 2) optimizer type, 3) learning rate, 4) weight decay, 5) embedding size, 6) linear layer size for Transformer or hidden size for RNN, and 7) dropout rate.

\section{Additional experiments with time-series classification}
\label{ucr_bench}

Additionally, we assessed the performance of our classification method on \textbf{UTS} (Univariate Time Series) using datasets obtained from the UCR archive ~\cite{UCR}, specifically the InsectSound and ElectricDevices datasets. Comprehensive descriptions of these datasets are publicly available through the UCR archive. 
We evaluate the performance of \SN on Univariate Time Series (UTS) classification against two datasets from the UCR archive ~\cite{UCR}. Our results, as displayed in Table~\ref{tab:ucr}, demonstrate that SeqNAS produces reasonable results for UTS classification due to its flexible search space. However, it should be noted that \SN performance suffers when used with small-size datasets from the UCR archive.

\begin{table}
  \caption{Comparison of UTS classification for datasets from USR archive with ROCKET, ROC-AUC is computed for both datasets.}
  \centering
  \label{tab:ucr}
\begin{tabular}{llllll}
    \toprule
    Dataset & SeqNAS & ROCKET \tablefootnote{Results are obtained from~~\cite{hydra}} & Train size & Test size  \\ 
    \midrule
    InsectSound & \bf{0.797} & 0.7823 & 25000 & 25000  \\
    ElectricDevices & 0.73 & \bf{0.7305} & 8926 & 7711  \\
    \bottomrule
\end{tabular}
\end{table}

\begin{table}
    \caption{Approximate search cost in \textbf{GPU hours} for SeqNAS, TextNAS and GRU with HPO. SeqNAS used 400 iterations. TextNAS uses ENAS for architecture search, which is significantly faster. For HPO details, we refer to Section~\ref{HPO}}
    \label{tab:costs}
    \centering
    \begin{tabular}{lccc}
    \toprule
    Dataset  &    SeqNas & TextNAS & GRU - HPO\\
    \midrule
    RBchurn &  60 &  1.2 & 2.1\\
    Taobao & 60 &  7.55 & 2.25\\
    AGE & 80 &  24.2 &  2.88\\
    \bottomrule
    \end{tabular}
\end{table}

\section{NAS-Bench Event Sequences}\label{sec:nas-bench-es}
Our dataset consists of 3200 architectures obtained on six different datasets. Out of the total architecture pool, 800 architectures were randomly queried from our search space, while 2400 architectures were queried using our search procedure. The distribution of these architectures across datasets and methods can be found in Table~\ref{tab:NAS BENCH}. For each architecture, we provide the best score achieved across all epochs, along with its feature vector encoded using \textbf{AVec}. The corresponding metrics for each score are listed in Table~\ref{tab:main_res}. All architectures within a dataset were trained for an equal number of epochs.

\begin{table}
    \caption{In this table, we demonstrate the distribution of architectures among various datasets and methods. The architectures are presented based on our final search procedure as well as randomly queried ones.}
    \centering
    \resizebox{0.5\textwidth}{!}{
    \begin{tabular}{llllllll}
       \toprule
         & Method / Dataset  & RBchurn  &  ABank  &  AmEx  & AGE & VBank & TaoBao  \\
         \midrule
         & Ours  &  400  &  400 & 400 &  400 & 400 & 400  \\
         & Random  &  400  &  400 & - &  - & - & - \\
         \bottomrule
    \end{tabular}}
    \label{tab:NAS BENCH}
\end{table}

\begin{figure}[ht]

\begin{center}
\begin{subfigure}[b]{0.4\linewidth}
\centering
\includegraphics[width=\linewidth]{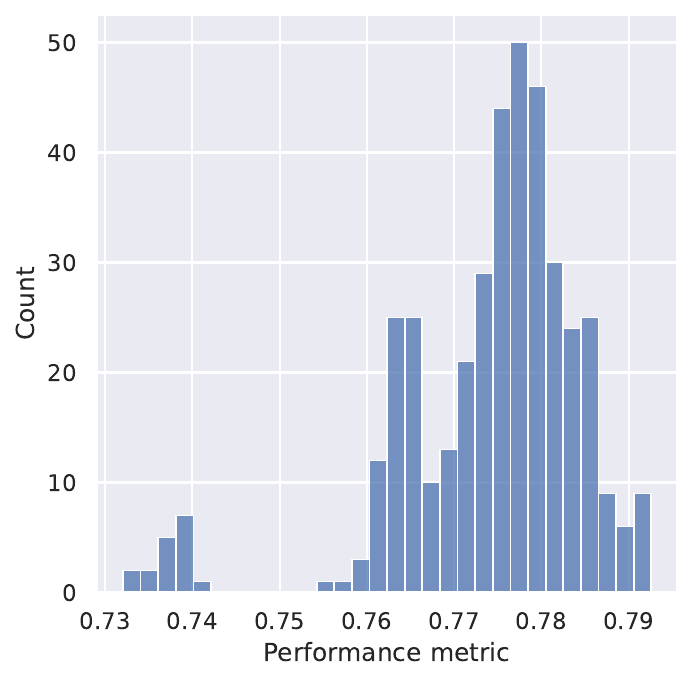}
\caption{ABank --- Random}
\end{subfigure}
\begin{subfigure}[b]{0.4\linewidth}
\centering
\includegraphics[width=\linewidth]{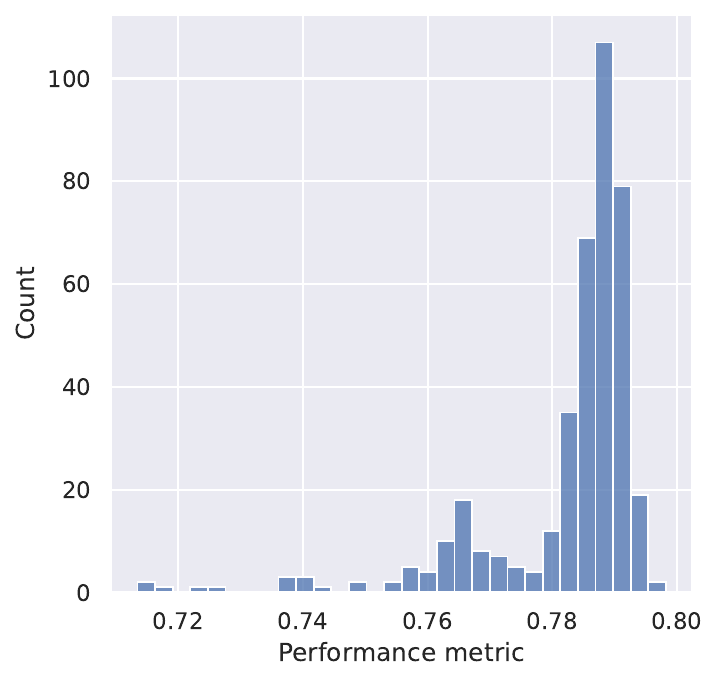}
\caption{ABank --- OUR}
\end{subfigure}

\begin{subfigure}[b]{0.4\linewidth}
\centering
\includegraphics[width=\linewidth]{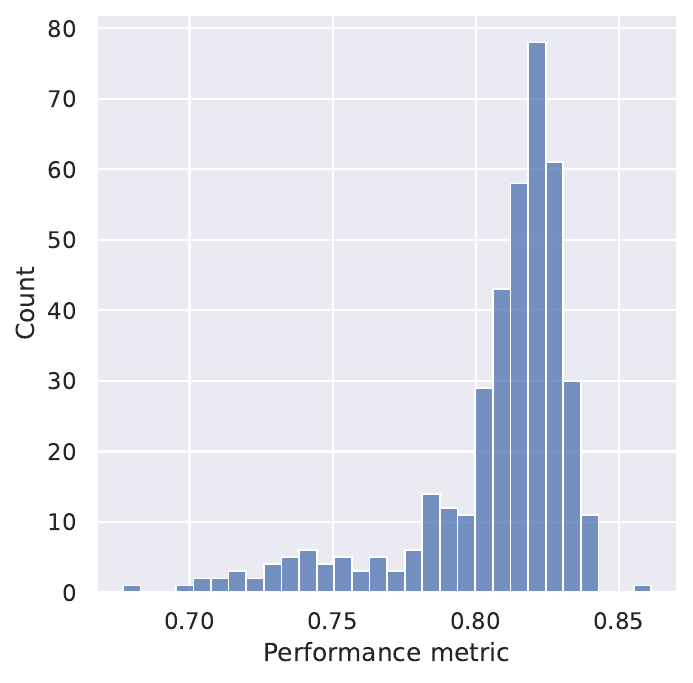}
\caption{RBchurn --- Random}
\end{subfigure}
\begin{subfigure}[b]{0.4\linewidth}
\centering
\includegraphics[width=\linewidth]{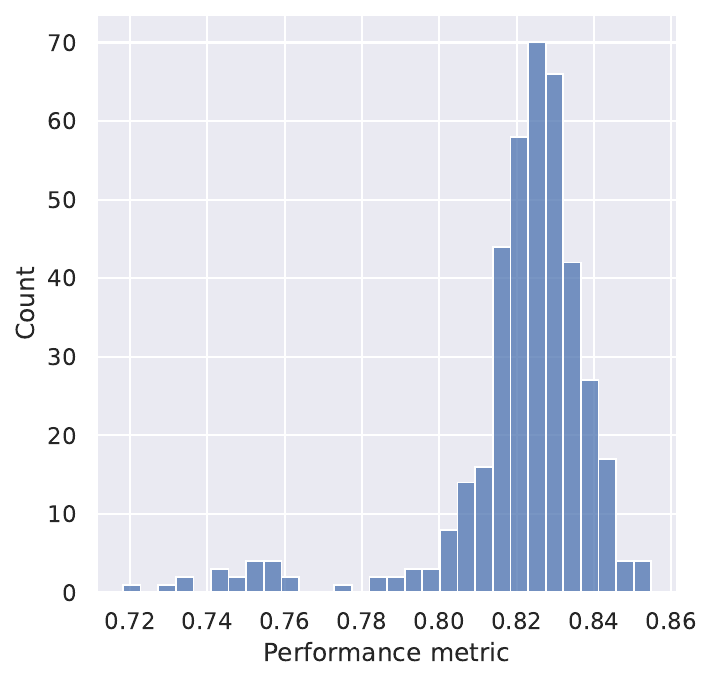}
\caption{RBchurn --- OUR}
\end{subfigure}
 
\caption{We present the distributions of queried architectures on two datasets using both our search procedure and random search. The distribution labeled with OUR represents our search procedure. It can be seen that architectures queried with OUR procedure are slightly shifted towards higher performance metrics, this shift is more significant for \textbf{ABank} dataset.}
\label{fig:data_hist}
\end{center}
\end{figure}

We specifically present architectures obtained through random search and our search procedure, as the architectures found through our procedure tend to have better performance metrics. This bias towards better performance can be observed in Figure~\ref{fig:data_hist}, which shows the distribution of queried architectures on two datasets using both our search procedure and random search.

\end{document}